\documentclass[letterpaper, 10 pt, conference]{ieeeconf}  % Comment this line out if you need a4paper

\IEEEoverridecommandlockouts                              % This command is only needed if
\usepackage{graphicx}
\usepackage{amsmath}
\usepackage{amssymb}
\usepackage{booktabs}
\usepackage{tabularx}
\usepackage{array}
\usepackage{url}
\usepackage{cite}
\usepackage{placeins}
\usepackage{xcolor}
\definecolor{revisionred}{RGB}{180,20,20}
\DeclareRobustCommand{\rev}[1]{{#1}}

\makeatletter
\let\ftype@table\ftype@figure
\makeatother

\title{\LARGE \bf
A Reconfigurable Bidirectional Cable-Driven Hip Exoskeleton With Swappable Bench/Backpack Dual-configuration Actuation
}

\author{Yuanlong Ji, \textit{Student Member, IEEE}, Shuhan Xiang, Qihan Ye,\\
and Xingbang Yang\textsuperscript{*}, \textit{Member, IEEE}%
\thanks{\textsuperscript{*}Research supported by the National Natural Science Foundation of China (Grant number 52475291), Beijing Natural Science Foundation (Grant number L222139 and QY26149), and the Fundamental Research Funds for the Central Universities (Grant numbers GW2025-71 and GW2026-MS-39). (Corresponding author: Xingbang Yang. E-mail: yangxingbang@buaa.edu.cn)}%
\thanks{Yuanlong Ji, Shuhan Xiang, Qihan Ye, and Xingbang Yang are with the School of Biological Science and Medical Engineering, Beihang University; the Key Laboratory of Biomechanics and Mechanobiology (Beihang University), Ministry of Education; Key Laboratory of Innovation and Transformation of Advanced Medical Devices, Ministry of Industry and Information Technology; National Medical Innovation Platform for Industry-Education Integration in Advanced Medical Devices (Interdiscipline of Medicine and Engineering), Beihang University, Beijing, 100191, China (E-mail: jiyuanlong@buaa.edu.cn).}%
}

\begin{document}
\raggedbottom
\setlength{\textfloatsep}{12pt plus 2pt minus 2pt}

\maketitle
\thispagestyle{empty}
\pagestyle{empty}

%%%%%%%%%%%%%%%%%%%%%%%%%%%%%%%%%%%%%%%%%%%%%%%%%%%%%%%%%%%%%%%%%%%%%%%%%%%%%%%%
\begin{abstract}
Hip exoskeletons provide an important hardware basis for lower-limb rehabilitation and locomotor assistance. \rev{Laboratory} rehabilitation assessment and system \rev{development} require substantial actuation and \rev{computing} resources, whereas mobile assistance \rev{requires} \rev{untethered} portability. \rev{Integrating} \rev{both} \rev{capabilities} within \rev{one reusable platform remains} a \rev{central} \rev{design challenge.} This paper presents a reconfigurable bidirectional cable-driven hip exoskeleton platform that rapidly switches between bench-mounted and backpack-mounted actuation while sharing one cable-free wearable hip interface. The platform modularly adapts the \rev{actuation} configuration, \rev{end-effector} sensing path, and low-level control interface. Each cable-driven end-effector weighs 0.405 kg, excluding the cable and \rev{actuation} unit, and integrates an encoder and a torque sensor; experiments validated bench-mounted \rev{admittance-based motion tracking} capability and backpack-mounted open-loop torque \rev{tracking.} Human-worn experiments with three healthy participants used myoMOTION to evaluate the platform's wearable-side hip-motion sensing capability, verified \rev{bench-to-backpack and backpack-to-bench} motion-ready switching across 30 trials \rev{in} \rev{$30.1\pm16.3$} s, and formed a small-scale multimodal wearable-exoskeleton gait dataset for sensing validation and data-driven algorithm development, comprising 8 min bench-mounted treadmill records and 11 min backpack-mounted outdoor walking records. These results show that, by unifying the wearable structure, actuation interface, and sensing path, the proposed platform enables validation of the same hip exoskeleton in both bench-mounted and backpack-mounted configurations, providing reusable hardware for iterative development and applications across scenarios.
% REVISION END B001
\end{abstract}

%%%%%%%%%%%%%%%%%%%%%%%%%%%%%%%%%%%%%%%%%%%%%%%%%%%%%%%%%%%%%%%%%%%%%%%%%%%%%%%%
\section{INTRODUCTION}

% ORIGINAL BEGIN B002 (baseline lines 74-74)
% The hip joint plays a central role in lower-limb mechanical work generation and postural regulation during locomotion \cite{Montgomery2018,Alexander2017}. Hip exoskeletons therefore provide an important hardware substrate for lower-limb rehabilitation, locomotor assistance, and biomechanics research \cite{Chen2020,Yang2022}. They have been used to improve walking economy in people after stroke \cite{Pruyn2026} and to support sit-to-stand transitions and walking efficiency in older adults \cite{Zhang2026}. They also serve as platforms for performance evaluation, control prototyping, and human motion data acquisition \cite{Ding2018,Kang2021}. These studies indicate that hip exoskeletons are not only assistive devices. They are also experimental platforms for evaluating assistance strategies, measuring human movement, and validating control methods.
% ORIGINAL END B002
% REVISION BEGIN B002
The hip joint plays a central role in lower-limb mechanical work generation and postural regulation during locomotion \cite{Montgomery2018,Alexander2017}. Hip exoskeletons therefore provide an important hardware substrate for lower-limb rehabilitation, locomotor assistance, and biomechanics research \cite{Chen2020,Yang2022}. They have been used to improve walking economy in people after stroke \cite{Pruyn2026} and to support sit-to-stand transitions and walking efficiency in older adults \cite{Zhang2026}. They also serve as platforms for performance evaluation, control prototyping, and human motion data acquisition \cite{Ding2018,Kang2021}. 
% REVISION END B002

% ORIGINAL BEGIN B003 (baseline lines 76-76)
% Existing hip and lower-limb exoskeletons are often integrated around a specific use case \cite{Karthik2025,Trott2026}. Bench-mounted or off-board actuation systems can relocate actuation, power supply, sensing acquisition, and computation away from the wearable side. This architecture provides larger actuation margins, stable power, richer interfaces, and programmable, repeatable assistance in controlled environments. It has therefore been widely used for assistance-parameter sweeping and multi-joint assistance comparison \cite{Bryan2021}, human-in-the-loop metabolic optimization and individualized assistance tuning \cite{Zhang2017,Ding2018,Witte2020}, and controlled rehabilitation interaction \cite{Wang2024,Kucuktabak2026}. However, these systems typically depend on fixed laboratory or clinical infrastructure and cannot be directly used for natural outdoor walking. In contrast, portable hip exoskeletons for locomotor assistance can keep the system mass low \cite{Pruyn2026,Zhang2026}. They are closer to real outdoor use, but must trade off mass, output capability, endurance, sensing interfaces, onboard computation, and safety redundancy \cite{Chen2020,Yang2022,Bajpai2024}.
% ORIGINAL END B003
% REVISION BEGIN B003
Existing hip and lower-limb exoskeletons are often integrated around a specific use case \cite{Karthik2025,Trott2026}. Bench-mounted or off-board actuation systems can relocate actuation, power supply, sensing acquisition, and computation away from the wearable \rev{hip interface.} This architecture provides larger actuation margins, stable power, richer interfaces, and programmable, repeatable assistance in controlled environments. It has therefore been widely used for assistance-parameter sweeping and multi-joint assistance comparison \cite{Bryan2021}, human-in-the-loop metabolic optimization and individualized assistance tuning \cite{Zhang2017,Ding2018,Witte2020}, and controlled rehabilitation interaction \cite{Wang2024,Kucuktabak2026}. However, \rev{their} \rev{external} \rev{equipment} \rev{and} \rev{tethered connections restrict mobility beyond} fixed laboratory or clinical \rev{infrastructure.} In contrast, portable hip exoskeletons \rev{carry their actuation and power supply in a lightweight system} for locomotor assistance \cite{Pruyn2026,Zhang2026}. They \rev{support} \rev{real-world} outdoor use, but must \rev{balance} output capability, endurance, sensing interfaces, onboard computation, and safety redundancy \rev{against carried mass} \cite{Chen2020,Yang2022,Bajpai2024}.
% REVISION END B003

% ORIGINAL BEGIN B004 (baseline lines 78-78)
% To expand the range of tasks supported by portable systems, recent work has pursued higher output capability and higher torque density under wearable constraints \cite{Bajpai2024}. Another route is modular lower-limb exoskeletons and open exoskeleton frameworks, which treat mechanical structures, sensing units, and control interfaces as configurable components \cite{ETHModular,Zhao2024,Williams2025}. Yet these two routes still leave a gap in cross-scenario continuity. High-output portable platforms increase actuation capacity in mobile settings, but onboard computation, sensing interfaces, and power space remain constrained. Modular frameworks improve structural reuse across tasks, but experiments across configurations still often require reassembly, rewiring, and low-level control adaptation \cite{Bajpai2024,Zhao2024,Williams2025}. As a result, bench-mounted test platforms and backpack-mounted assistive platforms commonly remain separate prototypes, which weakens continuity across development stages and application scenarios.
% ORIGINAL END B004
% REVISION BEGIN B004
To expand the range of tasks supported by portable systems, recent work has pursued higher output capability and higher torque density under wearable constraints \cite{Bajpai2024}. Another route is modular lower-limb exoskeletons and open exoskeleton frameworks, which treat mechanical structures, sensing units, and control interfaces as configurable components \cite{ETHModular,Zhao2024,Williams2025}. Yet these two routes still leave a gap in cross-scenario continuity. High-output portable platforms increase actuation capacity in mobile settings, but onboard computation, sensing interfaces, and power space remain constrained. Modular frameworks improve \rev{the} reuse \rev{of mechanical and control components} across tasks, \rev{while} \rev{laboratory} \rev{and} \rev{mobile operation} still require \rev{their} \rev{integration with the corresponding actuation} and \rev{sensing} \rev{hardware} \cite{Bajpai2024,Zhao2024,Williams2025}. As a result, bench-mounted test platforms and backpack-mounted assistive platforms commonly remain separate prototypes, \rev{limiting} \rev{reuse} \rev{of the same wearable structure and sensing arrangement} across development stages and application scenarios.
% REVISION END B004

% ORIGINAL BEGIN B005 (baseline lines 80-80)
% Decoupling the actuator from the human interface is a promising way to bridge this gap. Cable-driven and Bowden transmissions can place actuators away from the joint, reducing wearable-side mass and reflected inertia while preserving the advantages of hybrid rigid-soft structures for body conformity and force transmission \cite{Lee2017,Bryan2021}. This actuation form also allows one wearable interface, in principle, to accept drive sources with different locations and power levels without changing the main human-side load-bearing structure. However, Bowden cable transmissions introduce friction, hysteresis, pretension sensitivity, and time-varying cable configurations, which degrade the consistency between drive-side input and terminal output \cite{Dittli2021,Jammot2025,Chen2014,Wang2026}. Cross-scenario reuse therefore cannot rely only on relocating the actuator. The terminal sensing, cable interface, and low-level access path must also remain consistent across drive configurations.
% ORIGINAL END B005
% REVISION BEGIN B005
Decoupling the actuator from the human interface is a promising way to bridge this gap. Cable-driven and Bowden transmissions can place actuators away from the joint, reducing \rev{the} mass and \rev{inertia} \rev{of components carried on the moving limb} while preserving the advantages of hybrid rigid-soft structures for body conformity and force transmission \cite{Lee2017,Bryan2021}. This actuation form also allows one wearable interface, in principle, to accept \rev{actuation} sources with different locations and power levels without changing the main human-side load-bearing \rev{structure, thereby preserving the body attachments and joint alignment during actuation replacement.} However, Bowden cable transmissions introduce friction, hysteresis, pretension sensitivity, and time-varying cable configurations, which degrade the consistency between \rev{actuator-side} input and \rev{end-effector} output \cite{Dittli2021,Jammot2025,Chen2014,Wang2026}. Cross-scenario reuse therefore cannot rely only on relocating the actuator. The \rev{end-effector} \rev{sensing} \rev{arrangement} and \rev{mechanical} \rev{attachment} must \rev{be} \rev{compatible} \rev{with} \rev{both} \rev{configurations,} \rev{with corresponding sensor-acquisition and actuator-control connections.}
\rev{To enable laboratory and mobile experiments to reuse the same wearable structure and sensing arrangement, this} paper presents a reconfigurable bidirectional cable-driven hip exoskeleton platform \rev{that} \rev{decouples} the \rev{wearable} \rev{hip} \rev{interface} \rev{from} \rev{the} actuation \rev{units.} The platform uses a cable-free wearable hip interface as the common human-side structure. \rev{A} common \rev{quick-release/assembly} interface \rev{allows} the \rev{hip} \rev{exoskeleton} \rev{to} \rev{be driven interchangeably by either} bench-mounted \rev{or} backpack-mounted cable-driven \rev{units.} The contributions are threefold. First, we introduce a \rev{dual-configuration actuation} platform that \rev{supports} \rev{both resource-intensive} laboratory development and mobile-scenario validation \rev{through} a \rev{shared} human-side interface. Second, we design a \rev{quick-release/assembly} cable-driven end-effector that integrates torque sensing and angle measurement, and combine it with wireless inertial measurement units so that \rev{end-effector} torque, joint angle, and human motion \rev{retain} \rev{consistent} \rev{measurement types and locations} after \rev{actuation-unit} replacement. Third, we evaluate bench-mounted \rev{admittance-based motion tracking} response and the backpack-mounted open-loop torque \rev{tracking} in the current prototype, assess \rev{actuation} switching and continuous \rev{recording} \rev{of joint motion and interaction torque,} and use myoMOTION-verified human-worn records to establish a small-scale multimodal dataset with 8 min bench-mounted treadmill walking and 11 min backpack-mounted outdoor walking records. The dataset treats the exoskeleton as a wearable sensing form: signals are collected through the \rev{body} \rev{attachments} and \rev{end-effector} \rev{sensors} used \rev{during} actuation, providing \rev{time-aligned multimodal} data for end-to-end algorithms in multi-task human-exoskeleton interaction \cite{Zhang2026}.
% REVISION END B006

%%%%%%%%%%%%%%%%%%%%%%%%%%%%%%%%%%%%%%%%%%%%%%%%%%%%%%%%%%%%%%%%%%%%%%%%%%%%%%%%
\section{PLATFORM DESIGN}

\subsection{Task Requirements}

% ORIGINAL BEGIN B007 (baseline lines 89-89)
% To define the actuation requirements for the swappable drive paths, representative hip tasks were translated into peak-torque and speed targets. Sit-to-stand and functional training motivate a higher-margin bench-mounted path for controlled rehabilitation-oriented interaction \cite{Grimmer2026}, whereas level walking, ramps, and stair transitions motivate a lighter backpack-mounted path for mobile locomotion \cite{Camargo2021}. Accordingly, the bench-mounted configuration targets a 30 N\,m peak terminal response level and a 300 deg/s speed requirement, while the backpack-mounted configuration targets 10 N\,m and 150 deg/s. The corresponding component-level peak capabilities are 95.25 N\,m and approximately 360 deg/s for the bench-mounted drive path, and 53 N\,m and approximately 157 deg/s for the backpack-mounted path, consistent with Table~\ref{tab:comparison}. The resulting system requirement is to preserve the wearable hip interface, terminal sensing variables, and synchronized data stream while replacing the actuation source.
% ORIGINAL END B007
% REVISION BEGIN B007
To \rev{select} the \rev{two} actuation \rev{configurations,} representative \rev{hip-task} \rev{torques} were \rev{scaled} \rev{to} \rev{an 80 kg reference body mass. Selected sit-to-stand} and \rev{$18^\circ$} \rev{ramp-ascent} \rev{trajectories have peak hip torques of approximately 49.7} and \rev{60.1} \rev{N\,m,} \rev{respectively \cite{Grimmer2020,Camargo2021};} a \rev{selected} \rev{40\%} \rev{assistance} level \rev{corresponds} \rev{to 19.9--24.0 N\,m. For portable assistance, a selected 20\% level gives approximately 6.1} and \rev{5.6} N\,m \rev{for} \rev{1.00} \rev{m/s} level \rev{walking} and \rev{$5.2^\circ$} \rev{ramp ascent.} The corresponding \rev{speed} \rev{references} are \rev{approximately} \rev{298 deg/s for fast walking} and \rev{163--194} deg/s for the \rev{selected level and shallow-ramp tasks. To provide actuation margin for these loads and cable-transmission losses, the} bench-mounted \rev{and} \rev{backpack-mounted units use component-level peak torques of 95.25} and 53 N\,\rev{m, respectively, as detailed below. These task-dependent assistance levels guide component selection while the common wearable interface} and sensing \rev{arrangement} \rev{support} \rev{configuration} \rev{exchange.}
% REVISION END B007

\subsection{System Overview}

\begin{figure}[t]
    \centering
    \includegraphics[width=\linewidth]{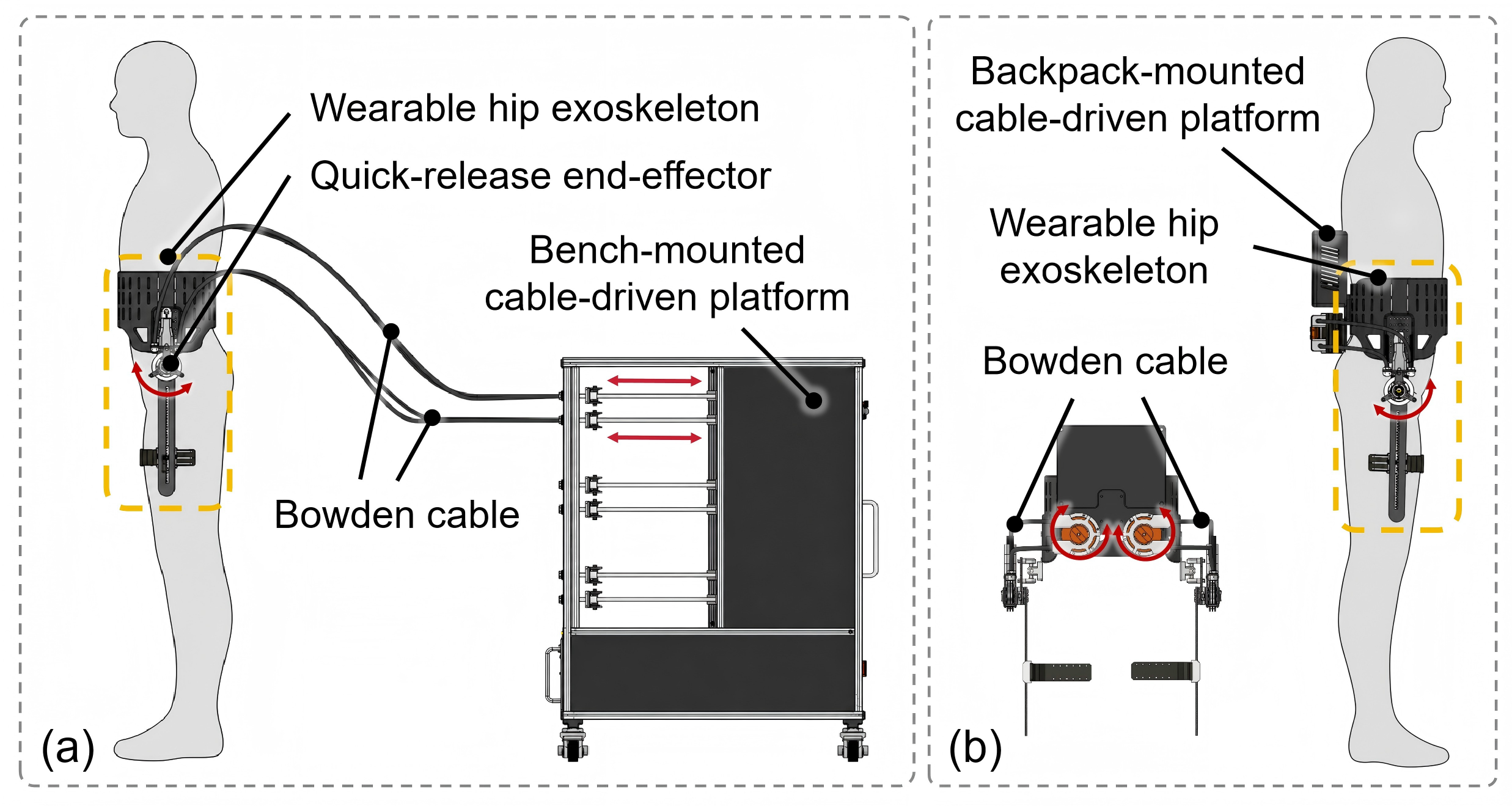}
    \caption{System configurations sharing the wearable hip interface. (a) Bench-mounted actuation. (b) Backpack-mounted actuation.}
    \label{fig:system_overview}
\end{figure}

\begin{figure*}[!t]
    \centering
    \includegraphics[width=0.95\textwidth]{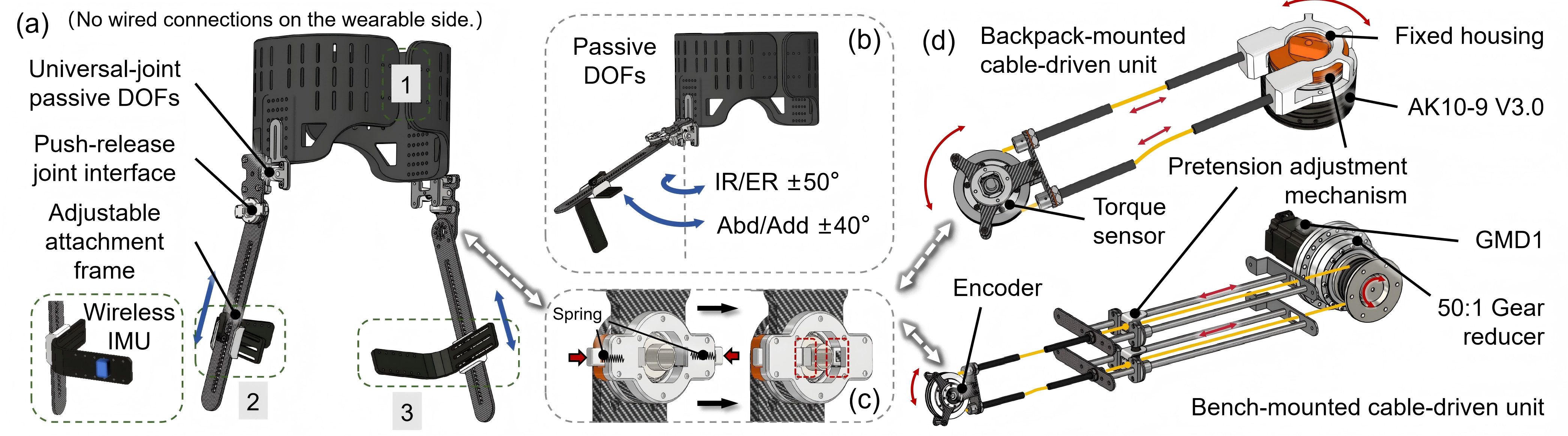}
% ORIGINAL BEGIN B009 (baseline lines 105-105)
%     \caption{Mechanical design for swappable actuation. (a) Wearable hip interface and wireless IMU placement. (b) Passive hip degrees of freedom. (c) Spring-loaded quick-release interface. (d) Backpack-mounted and bench-mounted bidirectional cable-driven units.}
% ORIGINAL END B009
% REVISION BEGIN B009
    \caption{Mechanical design for swappable actuation. (a) Wearable hip interface and wireless IMU placement. (b) Passive hip degrees of freedom. (c) Spring-loaded \rev{quick-release/assembly} interface. (d) Backpack-mounted and bench-mounted bidirectional cable-driven units.}
% REVISION END B009
    \label{fig:mechanism}
\end{figure*}

% ORIGINAL BEGIN B008 (baseline lines 100-100)
% The platform separates the wearable hip interface from the swappable actuation side, as summarized in Fig.~\ref{fig:system_overview}. The wearable side includes the waist and thigh attachments, passive hip degrees of freedom, quick-release sockets, and wireless inertial measurement units (IMUs). Each actuation unit has its own Bowden cables, motor-reducer modules, quick-release end-effectors, and computation/electrical path, but both units apply flexion-extension torque to the hip joint through the same quick-release interface. The bench-mounted configuration provides off-board actuation and high-resource laboratory access (Fig.~\ref{fig:system_overview}(a)), whereas the backpack-mounted configuration integrates the drive and lightweight control hardware on the waist-back side (Fig.~\ref{fig:system_overview}(b)). Fig.~\ref{fig:mechanism} shows the corresponding mechanical implementation.
% ORIGINAL END B008
% REVISION BEGIN B008
The platform separates the wearable hip interface from the swappable actuation \rev{units,} as summarized in Fig.~\ref{fig:system_overview}. The wearable \rev{hip interface} includes the waist and thigh attachments, passive hip \rev{joint} \rev{couplings,} \rev{quick-release/assembly} sockets, and wireless inertial measurement units (IMUs). Each actuation unit has its own Bowden cables, motor-reducer modules, \rev{quick-release/assembly} end-effectors, and \rev{computing} \rev{hardware and electrical connections,} but both units apply flexion-extension torque to the hip joint through the same \rev{quick-release/assembly} interface. The bench-mounted configuration provides off-board actuation and high-resource laboratory access (Fig.~\ref{fig:system_overview}(a)), whereas the backpack-mounted configuration integrates the \rev{actuation modules} and lightweight control hardware on the waist-back side (Fig.~\ref{fig:system_overview}(b)). Fig.~\ref{fig:mechanism} shows the corresponding mechanical implementation.
% REVISION END B008

% ORIGINAL BEGIN B010 (baseline lines 109-109)
% \subsection{Quick-Release Interface and Bidirectional Cable Routing}
% ORIGINAL END B010
% REVISION BEGIN B010
\subsection{\rev{Common Quick-Release/Assembly} Interface and Cable Routing}
% REVISION END B010

% ORIGINAL BEGIN B011 (baseline lines 111-111)
% The quick-release interface in Fig.~\ref{fig:mechanism}(c) consists of a wearable-side socket and an insertion component on the end-effector. Pressing the release component compresses the internal spring and disengages the lock; inserting a new end-effector lets the spring restore the locking component. This operation replaces the actuation unit without changing the waist or thigh structure. Bidirectional transmission is provided by flexion-side and extension-side cable paths that are pretensioned at the drive side. Each actuation unit uses a Bowden cable set matched to its layout, with a 3 mm-diameter Vectran cable core and a conduit consisting of an inner PTFE tube and an outer explosion-proof spiral tube. Both configurations keep a 1:1 relationship from the drive output shaft to the end-effector, preserving the output direction, angle definition, and terminal torque measurement target.
% ORIGINAL END B011
% REVISION BEGIN B011
The \rev{quick-release/assembly} interface in Fig.~\ref{fig:mechanism}(c) consists of a wearable-side socket and an insertion component on the end-effector. Pressing the release component compresses the internal spring and disengages the \rev{lock} \rev{for} \rev{removal.} \rev{After the replacement} end-effector \rev{is fully inserted, releasing the component allows} the spring \rev{to} restore the \rev{lock.} This operation replaces the actuation unit without changing the waist or thigh structure. Bidirectional transmission is provided by flexion-side and extension-side cable paths that are pretensioned at the \rev{actuator} side. Each actuation unit uses a Bowden cable set matched to its layout, with a 3 mm-diameter Vectran cable core and a conduit consisting of an inner PTFE tube and an outer \rev{spiral} \rev{protective} tube. Both configurations \rev{maintain} a 1:1 \rev{transmission ratio} from the \rev{actuator} output shaft to the end-effector, preserving the output direction, \rev{joint} angle \rev{convention,} and \rev{end-effector} torque measurement \rev{point.}
% REVISION END B011

\subsection{Bench-Mounted Cable-Driven Unit}

% ORIGINAL BEGIN B012 (baseline lines 115-115)
% The bench-mounted unit, shown in the lower part of Fig.~\ref{fig:mechanism}(d), places the motors, reducers, pretensioners, and mechanical stops on an external platform. It uses a GMD1-H401D30B-B060G1C motor (Googol Technology Co., Ltd., Shenzhen, China) and an LSS-32-XX-U-I reducer (Zhejiang Laifual Drive Co., Ltd., Shengzhou, China). The motor mass is 2.1 kg, the reduction ratio is 50:1, the component-level peak drive-side torque is 95.25 N\,m, and the output angular speed is approximately 360 deg/s. Together with external power, the motion-control card and high-performance industrial computer (Intel Core i7-12700 processor and NVIDIA GeForce RTX 4070 GPU) support high-bandwidth sensing acquisition, low-impedance interaction, and real-time algorithm debugging in laboratory experiments.
% ORIGINAL END B012
% REVISION BEGIN B012
The bench-mounted unit, shown in the lower part of Fig.~\ref{fig:mechanism}(d), places the motors, reducers, pretensioners, and mechanical stops on an external platform. It uses a GMD1-H401D30B-B060G1C motor (Googol Technology Co., Ltd., Shenzhen, China) and an LSS-32-XX-U-I reducer (Zhejiang Laifual Drive Co., Ltd., Shengzhou, China). The motor mass is 2.1 kg, the reduction ratio is 50:1, the component-level peak \rev{actuator-side} torque is 95.25 N\,m, and the output angular speed is approximately 360 deg/s. Together with external power, the motion-control card and high-performance industrial computer (Intel Core i7-12700 processor and NVIDIA GeForce RTX 4070 GPU) support high-bandwidth sensing acquisition, low-impedance interaction, and real-time algorithm debugging in laboratory experiments.
% REVISION END B012

\subsection{Backpack-Mounted Cable-Driven Unit}

% ORIGINAL BEGIN B013 (baseline lines 119-119)
% The backpack-mounted unit, shown in the upper part of Fig.~\ref{fig:mechanism}(d), integrates AK10-9 V3.0 drive modules (Nanchang Kude Intelligent Technology Co., Ltd., Nanchang, China), adjustable pretensioners, and low-level control hardware in a waist-back housing. Each AK10-9 V3.0 module weighs 0.94 kg, has a 9:1 reduction ratio, provides a component-level peak torque of 53 N\,m, and has an output angular speed of approximately 157 deg/s. A Jetson Nano embedded platform, an STM32 controller, and CAN-bus communication provide mobile recording, torque assistance, gait-recognition development, and outdoor experiment extensions while preserving the same quick-release interface and terminal sensing variables as the bench-mounted unit.
% ORIGINAL END B013
% REVISION BEGIN B013
The backpack-mounted unit, shown in the upper part of Fig.~\ref{fig:mechanism}(d), integrates AK10-9 V3.0 \rev{actuator} modules (Nanchang Kude Intelligent Technology Co., Ltd., Nanchang, China), adjustable pretensioners, and low-level control hardware in a waist-back housing. Each AK10-9 V3.0 module weighs 0.94 kg, has a 9:1 reduction ratio, provides a component-level peak torque of 53 N\,m, and has \rev{a no-load} output angular speed of \rev{1920} deg/s. A Jetson Nano embedded platform, an STM32 controller, and CAN-bus communication provide mobile recording, torque assistance, gait-recognition development, and outdoor experiment extensions while preserving the same \rev{quick-release/assembly} interface and \rev{end-effector} sensing \rev{arrangement} as the bench-mounted unit.
% REVISION END B013

\subsection{Wearable-Side Lightweight Sensing Integration}

% ORIGINAL BEGIN B014 (baseline lines 123-123)
% The wearable hip interface, shown in Fig.~\ref{fig:mechanism}(a), is shared by both actuation configurations. Waist and thigh attachments are adjusted to the participant, and a universal-joint-like passive hip connection accommodates approximately $\pm 50^\circ$ of internal-external rotation and $\pm 40^\circ$ of abduction-adduction motion (Fig.~\ref{fig:mechanism}(b)). The wearable side carries no wired electrical connections. Each quick-release end-effector weighs 0.405 kg, excluding the cable and drive unit, and integrates an M2210A2 torque sensor (Sunrise Instruments Co., Ltd., Nanning, China) and an MPT-1SB-25 C absolute encoder (Beijing KingKong Technology Co., Ltd., Beijing, China). Because both actuation units use the same type of end-effector and wireless IMU layout, terminal torque, terminal angle, and human motion data remain comparable after drive replacement.
% ORIGINAL END B014
% REVISION BEGIN B014
The wearable hip interface, shown in Fig.~\ref{fig:mechanism}(a), is shared by both actuation configurations. Waist and thigh attachments are adjusted to the participant, and a universal-joint-like passive hip connection accommodates approximately $\pm 50^\circ$ of internal-external rotation and $\pm 40^\circ$ of abduction-adduction motion (Fig.~\ref{fig:mechanism}(b)). The wearable \rev{hip interface} carries no wired electrical connections. Each \rev{quick-release/assembly} end-effector weighs 0.405 kg, excluding the cable and \rev{actuation} unit, and integrates an M2210A2 torque sensor (Sunrise Instruments Co., Ltd., Nanning, China) and an MPT-1SB-25 C absolute encoder (Beijing KingKong Technology Co., Ltd., Beijing, China). Because both actuation units \rev{share} the same \rev{end-effector} type and wireless IMU layout, \rev{end-effector} torque, \rev{joint} angle, and human motion \rev{measurements} remain comparable after \rev{actuation-unit} replacement.
% REVISION END B014

%%%%%%%%%%%%%%%%%%%%%%%%%%%%%%%%%%%%%%%%%%%%%%%%%%%%%%%%%%%%%%%%%%%%%%%%%%%%%%%%
\section{SWITCHABLE CONTROL}

\subsection{Common Sensing and Electrical Interface}

\begin{figure}[t]
    \centering
    \includegraphics[width=\linewidth]{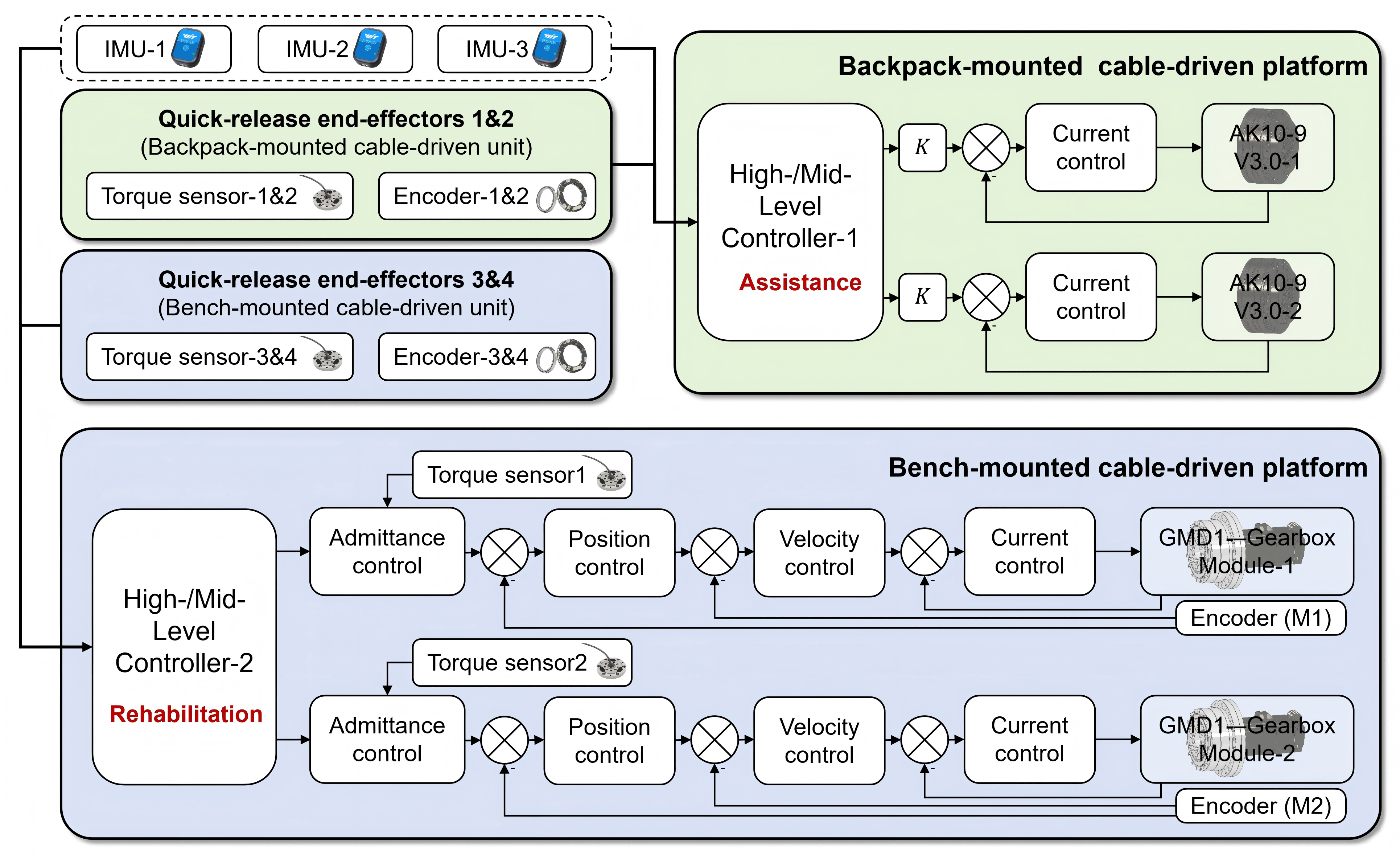}
% ORIGINAL BEGIN B017 (baseline lines 137-137)
%     \caption{Switchable control architecture for the wearable hip interface and terminal sensing inputs.}
% ORIGINAL END B017
% REVISION BEGIN B017
    \caption{Switchable control architecture for the wearable hip interface and \rev{end-effector} sensing inputs.}
% REVISION END B017
    \label{fig:control_architecture}
\end{figure}

% ORIGINAL BEGIN B015 (baseline lines 130-130)
% The control layer maps common sensing variables to configuration-dependent actuation paths. Hip angle $q$, terminal torque $\tau_s$, and wireless IMU data are used as shared inputs for control and recording. As shown in Fig.~\ref{fig:control_architecture}, the bench-mounted path uses terminal torque to generate a position command through an admittance model, whereas the backpack-mounted path converts a prescribed target torque into a drive input.
% ORIGINAL END B015
% REVISION BEGIN B015
The control layer maps common sensing variables to configuration-dependent \rev{control} paths. Hip angle $q$, \rev{end-effector} torque $\tau_s$, and wireless IMU data are \rev{available} \rev{through} \rev{common} \rev{sensing} \rev{interfaces.} \rev{End-effector torque supplies the bench-mounted admittance controller; in the backpack-mounted tests, torque, angle,} and \rev{IMU signals are recorded without end-effector feedback or a high-level assistance controller.} As shown in Fig.~\ref{fig:control_architecture}, the bench-mounted \rev{configuration} uses \rev{end-effector} torque to generate a position command through an admittance model, whereas the backpack-mounted \rev{configuration} converts a prescribed target torque into \rev{an} \rev{actuator} input.
% REVISION END B015

\begin{figure}[t]
    \centering
    \includegraphics[width=\linewidth]{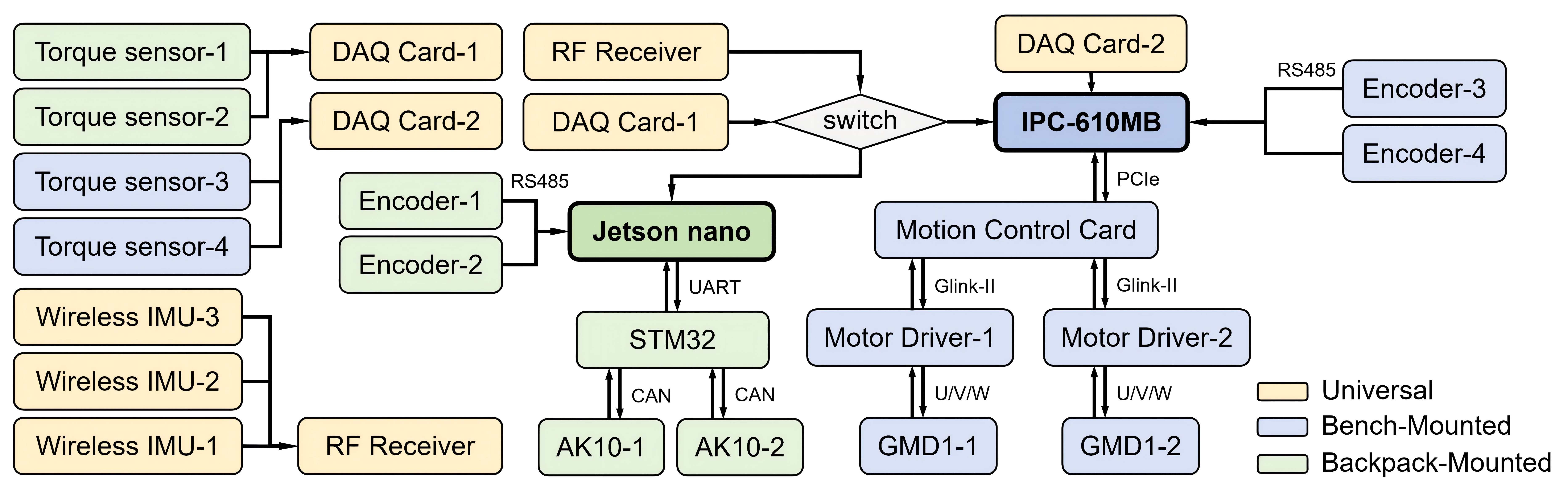}
    \caption{Common sensing and electrical adaptation for the two actuation configurations.}
    \label{fig:sensing_electrical}
\end{figure}

% ORIGINAL BEGIN B016 (baseline lines 132-132)
% Fig.~\ref{fig:sensing_electrical} summarizes the corresponding electrical access. The system includes three wireless IMUs, four torque sensors, and four encoders; the active terminal sensor pair is selected according to the installed drive unit. The bench-mounted path connects through data-acquisition and motion-control interfaces, whereas the backpack-mounted path connects through an embedded controller and CAN bus. Thus, the two configurations use different execution hardware while retaining common signal entry points.
% ORIGINAL END B016
% REVISION BEGIN B016
Fig.~\ref{fig:sensing_electrical} summarizes the corresponding \rev{sensor,} \rev{communication, and controller connections.} The system includes three wireless IMUs, four torque sensors, and four encoders; \rev{each} \rev{actuation} \rev{configuration} \rev{uses} \rev{its} \rev{own} \rev{two} \rev{torque} \rev{sensors} \rev{and} \rev{two} \rev{encoders.} The bench-mounted \rev{configuration} connects through data-acquisition and motion-control interfaces, whereas the backpack-mounted \rev{configuration} connects through an embedded controller and CAN bus. Thus, the two configurations use \rev{configuration-specific} \rev{computing and motor-control} hardware while retaining \rev{compatible} \rev{sensor} \rev{interfaces.}
% REVISION END B016

% ORIGINAL BEGIN B018 (baseline lines 148-148)
% \subsection{Bench-Mounted Admittance-Following Control}
% ORIGINAL END B018
% REVISION BEGIN B018
\subsection{Bench-Mounted \rev{Admittance-Based} \rev{Motion Tracking}}
% REVISION END B018

% ORIGINAL BEGIN B019 (baseline lines 150-150)
% The bench-mounted path uses position servoing as the low-level drive mode. For one hip joint, the terminal torque $\tau_s$ is offset by the static baseline $\tau_0$ and mapped to a position displacement $\Delta q$ through an admittance model. The desired position $q_d$ is then tracked by the position, velocity, and current loops of the bench-mounted drive.
% ORIGINAL END B019
% REVISION BEGIN B019
The bench-mounted \rev{configuration} uses position servoing as the low-level \rev{actuation} mode. For one hip joint, the \rev{end-effector} torque $\tau_s$ is offset by the static baseline $\tau_0$ and mapped to a position displacement $\Delta q$ through an admittance model. The desired position $q_d$ is then tracked by the position, velocity, and current loops of the bench-mounted \rev{actuator.}
% REVISION END B019

The admittance dynamics are written as
\begin{equation}
M_a \Delta \ddot q + B_a \Delta \dot q + K_a \Delta q
= \tau_s - \tau_0 .
\label{eq:admittance}
\end{equation}
% ORIGINAL BEGIN B020 (baseline lines 158-158)
% The desired position is then given by $q_d=q_0+\Delta q$, where $M_a$, $B_a$, and $K_a$ are the virtual mass, damping, and stiffness, respectively, and $q_0$ is the initial reference position. This formulation allows different interaction stiffness levels to be implemented through the same position-servo interface.
% ORIGINAL END B020
% REVISION BEGIN B020
The desired position is then given by $q_d=q_0+\Delta q$, where $M_a$, $B_a$, and $K_a$ are the virtual \rev{rotational inertia,} damping, and stiffness, respectively, and $q_0$ is the initial reference position. This formulation allows different \rev{inertial, damping, and} stiffness \rev{responses} to be implemented through the same position-servo interface.
% REVISION END B020

\subsection{Backpack-Mounted Torque Control}

% ORIGINAL BEGIN B021 (baseline lines 162-162)
% The backpack-mounted path uses open-loop torque control for bidirectional hip output. The target torque is defined as $\tau_d(t)=\gamma \tau_{\mathrm{ref}}(t)$, where $\tau_{\mathrm{ref}}(t)$ is the reference curve and $\gamma$ is the magnitude scaling factor; the sign of $\tau_d$ selects the flexion-side or extension-side cable. The controller converts $\tau_d$ into the drive input $u(t)$ using the torque coefficient $k_{\tau,p}$.
% ORIGINAL END B021
% REVISION BEGIN B021
The backpack-mounted \rev{configuration} uses open-loop torque control for bidirectional hip output. The target torque is defined as $\tau_d(t)=\gamma \tau_{\mathrm{ref}}(t)$, where $\tau_{\mathrm{ref}}(t)$ is the reference curve and $\gamma$ is the magnitude scaling factor; the sign of $\tau_d$ selects the flexion-side or extension-side cable. The controller converts $\tau_d$ into the \rev{actuator} input $u(t)$ using the torque coefficient $k_{\tau,p}$.
% REVISION END B021

The open-loop input mapping is
\begin{equation}
u(t)=\frac{\tau_d(t)}{k_{\tau,p}} .
\label{eq:torque_control}
\end{equation}
% ORIGINAL BEGIN B022 (baseline lines 169-169)
% Here, $u(t)$ is the drive input and $k_{\tau,p}$ is the nominal torque coefficient of the backpack-mounted drive. The nominal drive-side torque satisfies $\hat{\tau}(t)=k_{\tau,p}u(t)$. The terminal torque sensor measures $\tau_s$, allowing the command, cable transmission, and terminal response to be represented in the same torque space.
% ORIGINAL END B022
% REVISION BEGIN B022
Here, $u(t)$ is the \rev{actuator} input and $k_{\tau,p}$ is the nominal torque coefficient of the backpack-mounted \rev{actuator.} The nominal \rev{actuator-side} torque satisfies $\hat{\tau}(t)=k_{\tau,p}u(t)$. The \rev{end-effector} torque sensor measures $\tau_s$ \rev{independently, providing} the \rev{output measurement for evaluating torque tracking through the} cable \rev{transmission.}
% REVISION END B022

%%%%%%%%%%%%%%%%%%%%%%%%%%%%%%%%%%%%%%%%%%%%%%%%%%%%%%%%%%%%%%%%%%%%%%%%%%%%%%%%
\section{EXPERIMENTAL SETUP}

% ORIGINAL BEGIN B023 (baseline lines 174-174)
% The experiments were designed around three platform-level claims: the bench-mounted configuration can provide a low-impedance response through the admittance-following mode, the backpack-mounted configuration can generate measurable terminal torque under open-loop torque control, and drive replacement plus terminal sensing can support continuous use of the same wearable hip interface in controlled treadmill and natural outdoor walking. Human experiments were approved by the institutional ethics committee, and all participants provided written informed consent. Three healthy participants took part in the walking, switching, and comfort tests (age $25.0 \pm 1.7$ years, height $177.7 \pm 3.2$ cm, mass $72.7 \pm 6.4$ kg, mean $\pm$ SD). During human-worn tests, terminal angle, terminal torque, wireless IMU signals, and reference hip kinematics from a myoMOTION inertial motion capture system (Noraxon U.S.A. Inc., Scottsdale, AZ, USA) were recorded synchronously.
% ORIGINAL END B023
% REVISION BEGIN B023
The experiments were designed around three platform-level claims: the bench-mounted configuration can provide a low-impedance response through the \rev{admittance-based motion tracking} mode, the backpack-mounted configuration can generate measurable \rev{end-effector} torque under open-loop torque control, and \rev{actuation-unit} replacement plus \rev{end-effector} sensing can support continuous use of the same wearable hip interface in controlled treadmill and natural outdoor walking. Human experiments were approved by the institutional ethics committee, and all participants provided written informed consent. Three healthy participants took part in the walking, switching, and comfort tests (age $25.0 \pm 1.7$ years, height $177.7 \pm 3.2$ cm, mass $72.7 \pm 6.4$ \rev{kg).} During human-worn tests, \rev{end-effector} angle, \rev{end-effector} torque, wireless IMU signals, and hip kinematics \rev{measured by} a myoMOTION inertial motion capture system (Noraxon U.S.A. Inc., Scottsdale, AZ, USA) were recorded synchronously.
% REVISION END B023

\begin{figure*}[!t]
    \centering
    \includegraphics[width=0.98\textwidth]{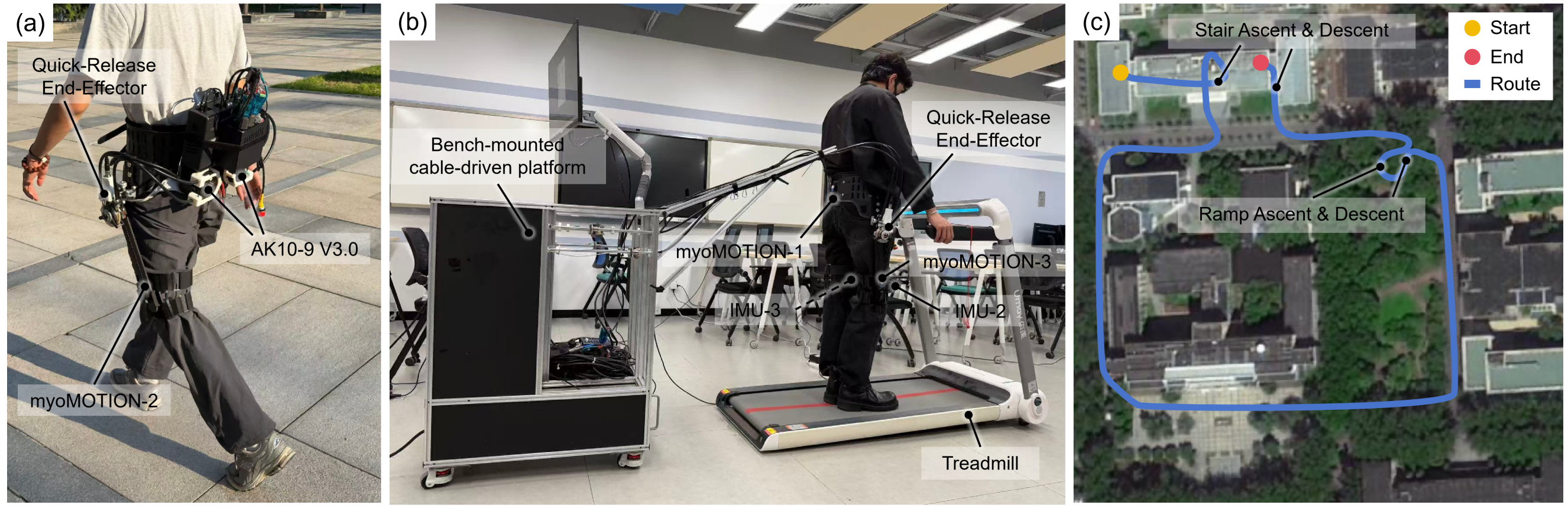}
    \caption{Human-worn data-collection setup. (a) Backpack-mounted outdoor walking configuration. (b) Bench-mounted treadmill walking configuration with myoMOTION reference sensors. (c) Outdoor route with start/end locations and terrain-transition segments.}
    \label{fig:human_setup}
\end{figure*}

\subsection{Low-Level Actuation Response Test}

% ORIGINAL BEGIN B024 (baseline lines 185-185)
% Low-level response tests were matched to the actual control entry point of each configuration. The bench-mounted configuration was evaluated in the human-worn admittance-following mode, using $M_a=0.05$ kg\,m$^2$, $B_a=0.05$ N\,m\,s/rad, and $K_a=0$ N\,m/rad. The low virtual mass, low damping, and zero stiffness reduce inertial, dissipative, and restoring terms, allowing human--device interaction torque to produce free-following hip motion through the position-servo cable path. The baseline-corrected terminal interaction torque $\tau_s-\tau_0$ was mapped through Eq.~\eqref{eq:admittance} and then tracked by the position-servo cable drive. The metrics were the agreement between encoder-based hip angle and the myoMOTION reference, together with the baseline-corrected terminal interaction torque in the same interval. This test evaluates whether terminal torque sensing, position-servo cable motion, and hip-motion sensing remain synchronized during human-worn walking.
% ORIGINAL END B024
% REVISION BEGIN B024
Low-level response tests were matched to the \rev{low-level} control \rev{mode} of each configuration. The bench-mounted configuration was evaluated in the human-worn \rev{admittance-based motion tracking} mode, using $M_a=0.05$ kg\,m$^2$, $B_a=0.05$ N\,m\,s/rad, and $K_a=0$ N\,m/rad. The low virtual \rev{rotational inertia,} low damping, and zero stiffness reduce inertial, dissipative, and restoring terms, allowing human--device interaction torque to produce free-following hip motion through the position-servo cable path. The baseline-corrected \rev{end-effector} interaction torque $\tau_s-\tau_0$ was mapped through Eq.~\eqref{eq:admittance} and then tracked by the position-servo cable \rev{actuator.} The metrics were the agreement between encoder-based hip angle and the myoMOTION reference, together with the baseline-corrected \rev{end-effector} interaction torque in the same interval. This test evaluates whether \rev{end-effector} torque sensing, position-servo cable motion, and hip-motion sensing remain synchronized during human-worn walking.
% REVISION END B024

% ORIGINAL BEGIN B025 (baseline lines 187-187)
% The backpack-mounted configuration was evaluated under open-loop torque control. The ramp-hold command rose to 10 N\,m at 0.300 N\,m/s and held near the target for 10 s; the unloading rate was 0.500 N\,m/s. The bidirectional sinusoidal command had a 5 N\,m amplitude at 0.25 Hz, and each window contained 10 cycles. For the backpack-mounted data reported here, five ramp-output windows and five complete sinusoidal windows were retained, and holding statistics were computed only from windows containing a complete hold segment. For open-loop torque tests, the terminal response ratio was defined as
% ORIGINAL END B025
% REVISION BEGIN B025
The backpack-mounted configuration was evaluated under open-loop torque control. The ramp-hold command rose to 10 N\,m at 0.300 N\,m/s and held near the target for 10 s; the unloading rate was 0.500 N\,m/s. The bidirectional sinusoidal command had a 5 N\,m amplitude at 0.25 Hz, and each window contained 10 cycles. \rev{The 5--10 N\,m input range was selected for basic low-load hip-exoskeleton operation; the 0.25 Hz sinusoid evaluated repeated bidirectional loading, while the ramp-hold input evaluated loading and sustained output.} For the backpack-mounted data reported here, five \rev{complete} \rev{ramp-hold sequences} and five complete sinusoidal windows were \rev{retained; each sinusoidal window contained ten cycles,} and \rev{all} \rev{five} \rev{ramp} \rev{sequences} \rev{included} a complete hold segment. For open-loop torque tests, the \rev{end-effector} response ratio was defined as
% REVISION END B025
\begin{equation}
\eta_\tau =
\frac{\tau_s}{\tau_m}.
\label{eq:terminal_response_ratio}
\end{equation}
% ORIGINAL BEGIN B026 (baseline lines 193-193)
% Here, $\tau_m=k_{\tau,p}^{nom}u$ is the nominal drive-side torque of the backpack-mounted unit, $\tau_s$ is the terminal torque measured by the end-effector sensor, and $k_{\tau,p}^{nom}=1.68$ N\,m/A is the nominal torque coefficient of the backpack-mounted drive. This ratio summarizes how the drive-side input appears at the end-effector under the selected cable path and pretension.
% ORIGINAL END B026
% REVISION BEGIN B026
Here, $\tau_m=k_{\tau,p}^{nom}u$ is the nominal \rev{actuator-side} torque of the backpack-mounted unit, $\tau_s$ is the \rev{end-effector} torque measured by the end-effector sensor, and $k_{\tau,p}^{nom}=1.68$ N\,m/A is the nominal \rev{output-torque/current} \rev{ratio calculated from the rated values} of \rev{18} \rev{N\,m} \rev{and 10.7 A.} This ratio summarizes \rev{torque} \rev{transmission} under the selected cable path and pretension. \rev{The ramp-up value is the slope of a zero-intercept fit over $2\leq\tau_m\leq9.5$ N\,m; the holding value is the mean measured-to-commanded torque ratio during the plateau, and the sinusoidal value is the ratio of fitted fundamental amplitudes.}
% REVISION END B026

% ORIGINAL BEGIN B027 (baseline lines 195-195)
% \subsection{Drive Switching and Comfort Test}
% ORIGINAL END B027
% REVISION BEGIN B027
\subsection{\rev{Actuation} Switching and Comfort Test}
% REVISION END B027

% ORIGINAL BEGIN B028 (baseline lines 197-197)
% The switching test measured recovery to a motion-ready state, rather than mechanical detachment alone. With the participant standing under experimenter protection, each participant completed five bench-to-backpack and five backpack-to-bench switches, yielding 30 records. The timing window included end-effector release and locking, cable-tension checking, sensing online confirmation, and low-level control enabling. Motion-ready status was confirmed by normal sensor streaming, drive online feedback, and low-level control enablement. Overall comfort was rated before and immediately after each walking condition on a 1--7 scale adapted from \cite{Bryan2021}.
% ORIGINAL END B028
% REVISION BEGIN B028
The switching test measured \rev{the} recovery \rev{time} to a motion-ready \rev{state} \rev{from} \rev{one} \rev{actuation} \rev{configuration} \rev{to the other.} With the participant standing under experimenter protection, each participant completed five bench-to-backpack and five backpack-to-bench switches, yielding 30 records. The timing window included end-effector release and locking, cable-tension checking, sensing online confirmation, and low-level control enabling. Motion-ready status was confirmed by normal sensor streaming, \rev{actuator} \rev{status} feedback, and low-level control enablement. Overall comfort was rated before and immediately after each walking condition on a 1--7 scale adapted from \cite{Bryan2021}.
% REVISION END B028

\subsection{Human-Worn Data Collection}

% ORIGINAL BEGIN B029 (baseline lines 201-201)
% Each participant completed two data-collection sessions: an 8 min bench-mounted treadmill walk at 1.0 m/s and a backpack-mounted outdoor walk without active assistance. The bench-mounted session used the same admittance-following mode described above. The outdoor route included level walking, ramps, stairs, turns, start--stop transitions, and self-selected speed changes (Fig.~\ref{fig:human_setup}(c)). The duration provided sustained walking records for synchronization and gait-variability analysis \cite{ATS2002,Konig2014}. The backpack-mounted and bench-mounted wearing setups and sensor placement are shown in Fig.~\ref{fig:human_setup}(a) and (b), respectively; a supplementary video documents one representative workflow.
% ORIGINAL END B029
% REVISION BEGIN B029
Each participant completed two data-collection sessions: an 8 min bench-mounted treadmill walk at 1.0 m/s and a backpack-mounted outdoor walk without active assistance. The \rev{outdoor sessions lasted 12.68, 11.06, and 9.76 min, respectively ($11.17\pm1.47$ min). The} bench-mounted session used the same \rev{admittance-based motion tracking} mode described above. \rev{All participants followed the same} outdoor \rev{route,} \rev{including} level walking, ramps, stairs, turns, start--stop transitions, and self-selected speed changes (Fig.~\ref{fig:human_setup}(c)). The duration provided sustained walking records for synchronization and gait-variability analysis \cite{ATS2002,Konig2014}. The backpack-mounted and bench-mounted wearing setups and \rev{the} placement \rev{of myoMOTION sensors and wireless IMUs} are shown in Fig.~\ref{fig:human_setup}(a) and (b), \rev{respectively.}
% REVISION END B029

For dataset analysis, encoder-based hip angles were aligned to the myoMOTION reference and evaluated using calibrated RMSE, MAE, and correlation coefficient. Channel-level acquisition quality was summarized by valid-frame ratios and effective update rates. For each channel $c$,
\begin{equation}
R_c=\frac{N_{c,\mathrm{valid}}}{f_{\log}T}\times 100\%, \quad
f_{c,\mathrm{eff}}=\frac{N_{c,\mathrm{update}}}{T},
\end{equation}
% ORIGINAL BEGIN B030 (baseline lines 208-208)
% where $T$ is the recording duration, $f_{\log}=100$ Hz for the processed bench-mounted records, $N_{c,\mathrm{valid}}$ is the number of valid logging frames, and $N_{c,\mathrm{update}}$ is the number of non-repeated terminal samples or unique wireless-IMU packets. Apparent terminal resistance during walking was summarized by subtracting the median torque in the first 5 s static window from each terminal torque channel and averaging the absolute baseline-corrected torque across the two channels. The synchronized recordings were organized as a small-scale multimodal wearable-exoskeleton gait dataset containing anonymized raw data, aligned files, a data dictionary, processing scripts, and an anonymized outdoor-route description. This organization supports feasibility assessment of the exoskeleton as a wearable sensing platform and provides paired inputs for end-to-end data-driven gait and interaction algorithms.
% ORIGINAL END B030
% REVISION BEGIN B030
where $T$ is the recording duration, $f_{\log}=100$ Hz \rev{is} the \rev{configured} \rev{logging} \rev{rate for both configurations,} $N_{c,\mathrm{valid}}$ is the number of logging \rev{frames with finite numerical values in channel $c$,} and $N_{c,\mathrm{update}}$ is the number of non-repeated \rev{end-effector} samples or unique wireless-IMU packets. \rev{These effective update rates characterize the recorded data streams.} Apparent \rev{end-effector} resistance during walking was summarized by subtracting the \rev{channel-specific} median torque \rev{over} the first 5 s \rev{of} each \rev{aligned} \rev{record} and averaging the absolute baseline-corrected torque \rev{over the remaining samples and} across the two channels. The synchronized recordings were organized as a small-scale multimodal wearable-exoskeleton gait dataset containing anonymized raw data, aligned files, a data dictionary, processing scripts, and an anonymized outdoor-route description. This organization supports feasibility assessment of the exoskeleton as a wearable sensing platform and provides paired inputs for end-to-end data-driven gait and interaction algorithms.
% REVISION END B030

%%%%%%%%%%%%%%%%%%%%%%%%%%%%%%%%%%%%%%%%%%%%%%%%%%%%%%%%%%%%%%%%%%%%%%%%%%%%%%%%
\section{EXPERIMENTAL RESULTS}

\subsection{Low-Level Actuation Response}

\begin{figure}[t]
    \centering
    \includegraphics[width=0.90\linewidth]{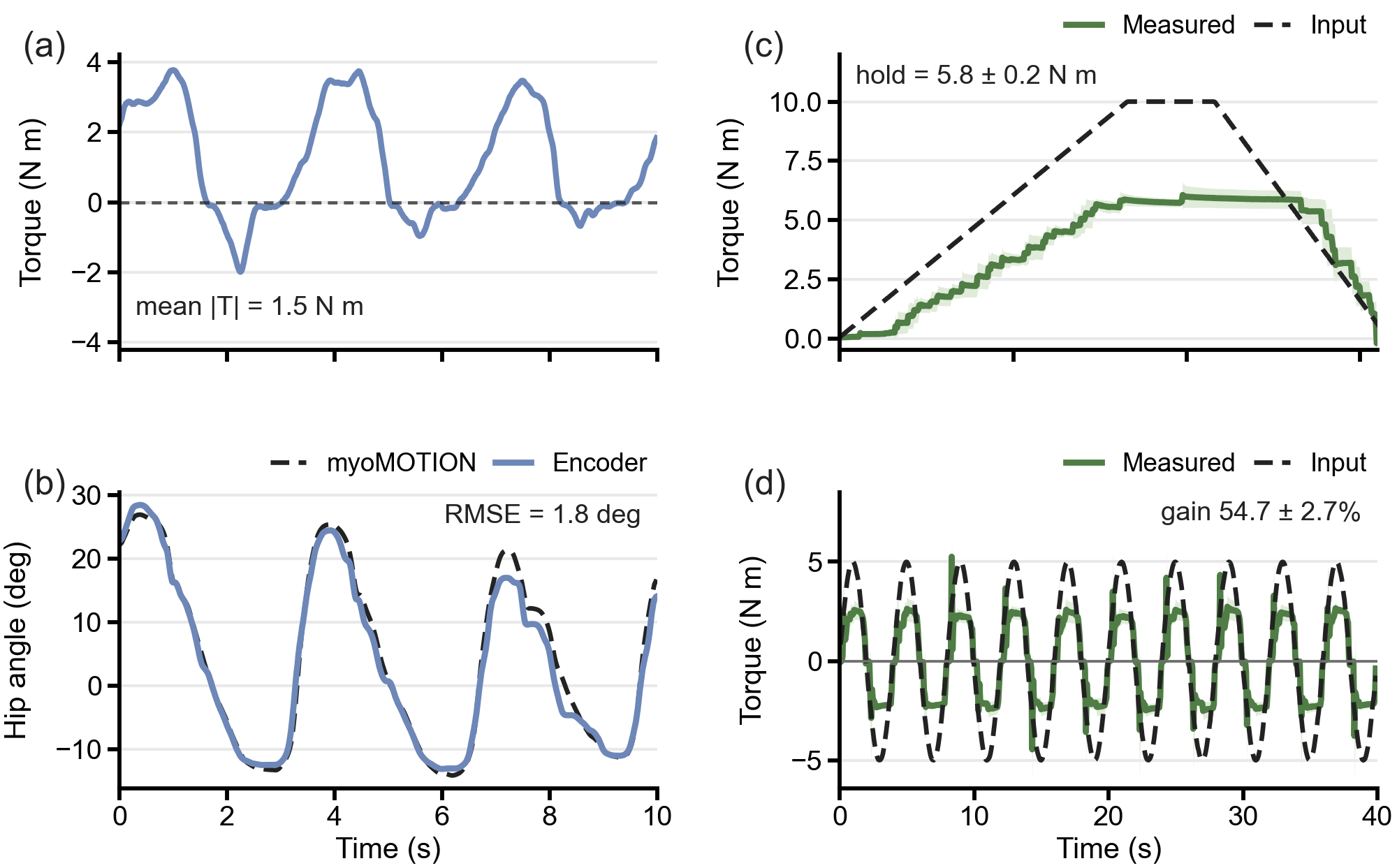}
% ORIGINAL BEGIN B031 (baseline lines 218-218)
%     \caption{Low-level actuation responses of the two configurations. (a) Terminal interaction torque during bench-mounted admittance-following walking. (b) Time-aligned hip angles from myoMOTION and the terminal encoder. (c) Backpack-mounted open-loop response to a 10 N\,m ramp-hold input. (d) Backpack-mounted open-loop response to a $\pm 5$ N\,m, 0.25 Hz bidirectional sinusoidal input.}
% ORIGINAL END B031
% REVISION BEGIN B031
    \caption{Low-level actuation responses of the two configurations. (a) \rev{End-Effector} interaction torque during bench-mounted \rev{admittance-based motion tracking} walking. (b) Time-aligned hip angles from myoMOTION and the \rev{end-effector} encoder. (c) Backpack-mounted open-loop response to a 10 N\,m ramp-hold input. (d) Backpack-mounted open-loop response to a $\pm 5$ N\,m, 0.25 Hz bidirectional sinusoidal input.}
% REVISION END B031
    \label{fig:static_output}
\end{figure}

% ORIGINAL BEGIN B034 (baseline lines 236-236)
% The low-level response results were interpreted according to the control mode used by each actuation configuration. In the bench-mounted admittance-following condition, the mean absolute baseline-corrected terminal interaction torque was 1.48 N\,m; over the same time window, the encoder-based hip angle preserved the periodic pattern of the myoMOTION reference, with an RMSE of $1.80^\circ$ and a correlation coefficient of 0.994 (Fig.~\ref{fig:static_output}(a,b)). This time-aligned relationship indicates that the admittance-following mode enabled continuous hip following in response to human--device interaction torque while maintaining low apparent interaction torque. For the completed backpack-mounted tests, the 10 N\,m ramp command produced a repeatable terminal loading response (Fig.~\ref{fig:static_output}(c)). Across five complete ramp-up windows, the rise-fit response ratio was $56.7 \pm 1.7\%$; during the holding phase, the terminal torque was $5.84 \pm 0.20$ N\,m and the holding response ratio was $58.4 \pm 2.0\%$. Under the $\pm 5$ N\,m, 0.25 Hz sinusoidal input, the fitted terminal amplitude was $2.74 \pm 0.13$ N\,m, corresponding to an amplitude ratio of $54.7 \pm 2.7\%$ and a phase lag of $10.4 \pm 0.7^\circ$ (Fig.~\ref{fig:static_output}(d)). Together, the ramp and sinusoidal tests show that the backpack-mounted cable path produced repeatable end-effector responses under the selected pretension and quick-release interface.
% ORIGINAL END B034
% REVISION BEGIN B034
The low-level response results were interpreted according to the control mode used by each actuation configuration. In the bench-mounted \rev{admittance-based motion tracking} condition, the mean absolute baseline-corrected \rev{end-effector} interaction torque was 1.48 N\,m; over the same time window, the encoder-based hip angle preserved the periodic pattern of the myoMOTION reference, with an RMSE of $1.80^\circ$ and a correlation coefficient of 0.994 (Fig.~\ref{fig:static_output}(a,b)). This time-aligned relationship indicates that the \rev{admittance-based motion tracking} mode enabled continuous hip following in response to human--device interaction torque while maintaining low apparent interaction torque. For the completed backpack-mounted tests, the 10 N\,m ramp command produced a repeatable \rev{end-effector} loading response (Fig.~\ref{fig:static_output}(c)). Across five complete ramp-up windows, the rise-fit response ratio was $56.7 \pm 1.7\%$; during the holding phase, the \rev{end-effector} torque was $5.84 \pm 0.20$ N\,m and the holding response ratio was $58.4 \pm 2.0\%$. Under the $\pm 5$ N\,m, 0.25 Hz sinusoidal input, the fitted \rev{end-effector} amplitude was $2.74 \pm 0.13$ N\,m, corresponding to an amplitude ratio of $54.7 \pm 2.7\%$ and a phase lag of $10.4 \pm 0.7^\circ$ (Fig.~\ref{fig:static_output}(d)). Together, the ramp and sinusoidal tests show that the backpack-mounted cable path produced repeatable end-effector responses under the selected pretension and \rev{quick-release/assembly} interface.
% REVISION END B034

% ORIGINAL BEGIN B035 (baseline lines 238-238)
% \subsection{Drive Switching and Comfort}
% ORIGINAL END B035
% REVISION BEGIN B035
\subsection{\rev{Actuation} Switching and Comfort}

\begin{figure}[t]
        \centering
        \includegraphics[width=\linewidth]{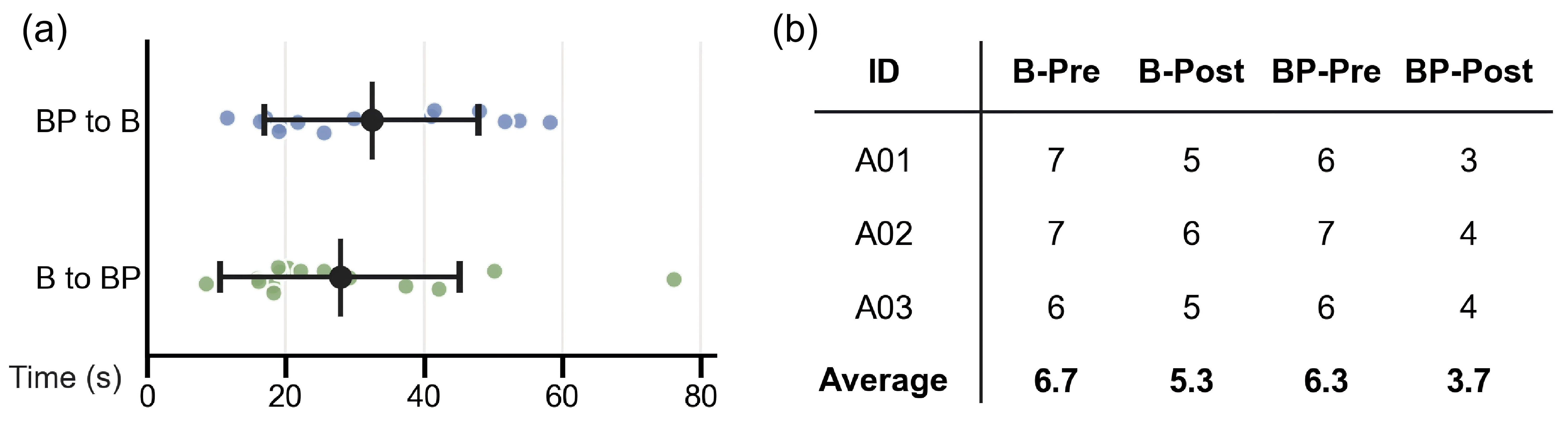}
% ORIGINAL BEGIN B032 (baseline lines 225-225)
%         \caption{Drive switching and comfort results. (a) Time to motion-ready state for bench-to-backpack and backpack-to-bench switches. Points denote individual trials, and error bars indicate mean $\pm$ SD. (b) Overall comfort scores collected before and immediately after the corresponding walking condition. B and BP denote the bench-mounted and backpack-mounted configurations, respectively. Scores range from 1 to 7, with higher values indicating better comfort.}
% ORIGINAL END B032
% REVISION BEGIN B032
        \caption{\rev{Actuation} switching and comfort results. (a) Time \rev{required} to \rev{restore} \rev{motion} \rev{readiness} \rev{after changing between the bench-mounted} and \rev{backpack-mounted} \rev{configurations.} Points denote individual trials, and error bars indicate mean $\pm$ SD. (b) \rev{Participant-reported overall} comfort scores before and immediately after the corresponding walking condition. B and BP denote the bench-mounted and backpack-mounted configurations, respectively. Scores range from 1 to 7, with higher values indicating better comfort.}
% REVISION END B032
        \label{fig:switching}
\end{figure}
% REVISION END B035

% ORIGINAL BEGIN B036 (baseline lines 240-240)
% All switching trials recovered a motion-ready system while keeping the wearable hip interface unchanged. Based on the 30 raw switching records, the overall switching time was $30.1 \pm 16.3$ s, with a median of 23.8 s. Fig.~\ref{fig:switching}(a) shows no clear directional bias between bench-to-backpack and backpack-to-bench switching; the longer tail is primarily associated with manual end-effector locking and cable-tension confirmation. Fig.~\ref{fig:switching}(b) shows that comfort decreased from 6.7 to 5.3 after bench-mounted treadmill walking and from 6.3 to 3.7 after backpack-mounted outdoor walking, indicating that carried load and terrain transitions dominate the subjective burden in the mobile condition.
% ORIGINAL END B036
% REVISION BEGIN B036
All switching trials recovered a motion-ready system while keeping the wearable hip interface unchanged. Based on the 30 raw switching records, the overall switching time was $30.1 \pm 16.3$ s, with a median of 23.8 s. Fig.~\ref{fig:switching}(a) shows no clear directional bias between bench-to-backpack and backpack-to-bench switching; \rev{experimenter observations associated} the longer \rev{switching} \rev{times} with manual end-effector locking and cable-tension confirmation. Fig.~\ref{fig:switching}(b) shows that comfort decreased from 6.7 to 5.3 after bench-mounted treadmill walking and from 6.3 to 3.7 after backpack-mounted outdoor walking, indicating that carried load and terrain transitions dominate the subjective burden in the mobile condition.
% REVISION END B036

\subsection{Human-Worn Data Recording}

\begin{figure}[t]
        \centering
        \includegraphics[width=\linewidth]{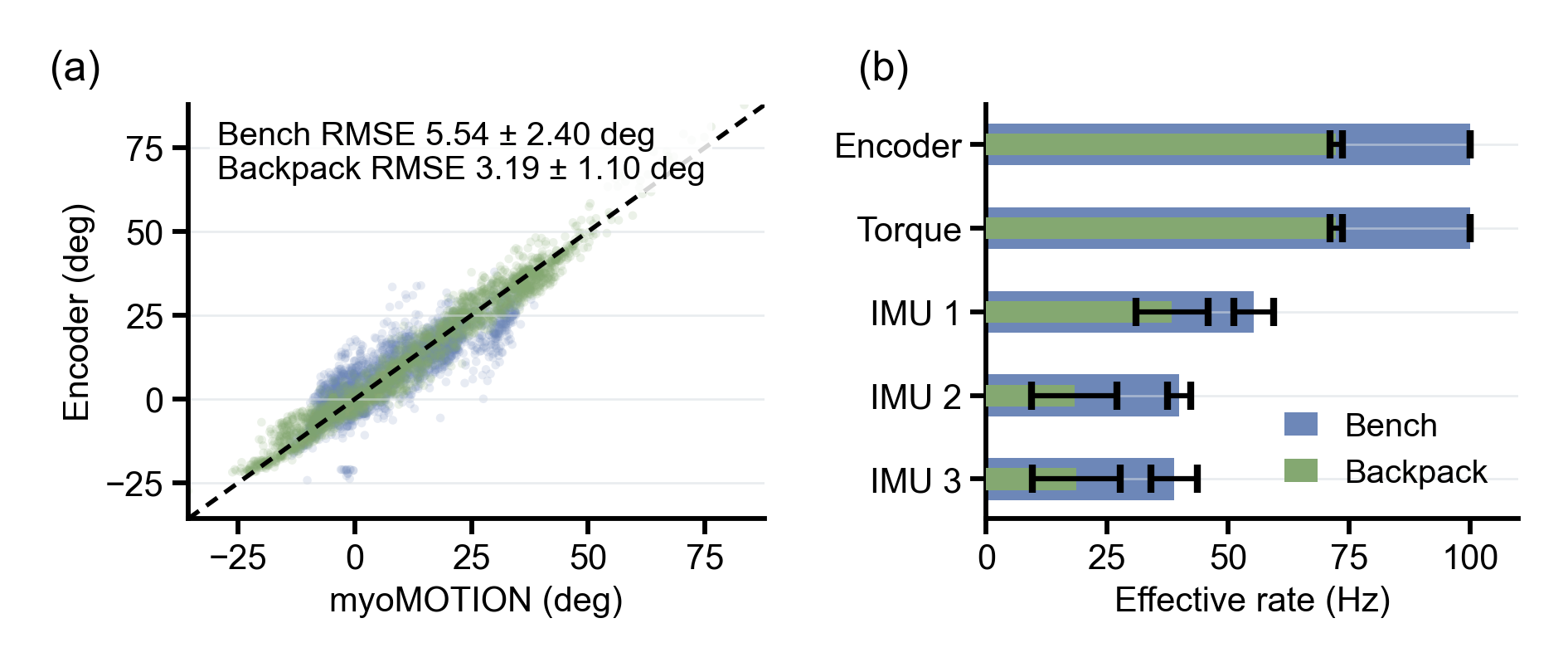}
% ORIGINAL BEGIN B033 (baseline lines 232-232)
%         \caption{Human-worn data recording results. (a) Sample-wise agreement between encoder-based hip angle and the myoMOTION reference pooled across the bench-mounted and backpack-mounted records. (b) Effective update rates of terminal sensing and wireless IMU channels, with backpack-mounted outdoor records overlaid in green.}
% ORIGINAL END B033
% REVISION BEGIN B033
        \caption{Human-worn data recording results. (a) Sample-wise agreement between encoder-based hip angle and the myoMOTION reference pooled across the bench-mounted and backpack-mounted records. (b) Effective update rates of \rev{end-effector} sensing and wireless IMU channels, with backpack-mounted outdoor records overlaid in green.}
% REVISION END B033
        \label{fig:human_recording}
\end{figure}

% ORIGINAL BEGIN B037 (baseline lines 244-244)
% The human-worn records show that the shared terminal sensing chain preserves the main hip-motion pattern in both use conditions. In Fig.~\ref{fig:human_recording}(a), encoder-based hip angles remain aligned with the myoMOTION reference, with calibrated RMSE values of $5.54 \pm 2.40^\circ$ for bench-mounted walking and $3.19 \pm 1.10^\circ$ for backpack-mounted walking; the residual spread reflects alignment, soft-tissue motion, and wearable-interface compliance. After static-baseline subtraction, the mean absolute terminal torque was $1.05 \pm 0.32$ N\,m in the bench-mounted admittance-following condition and $0.68 \pm 0.35$ N\,m in the backpack-mounted unpowered condition, indicating low apparent resistance in both walking modes. This low-resistance condition is useful for intention-sensing extensions and wearable data collection, because terminal measurements can remain engaged while adding limited mechanical burden. Fig.~\ref{fig:human_recording}(b) further indicates that terminal encoder and torque channels provide stable high-rate records, whereas the wireless IMU channel is more variable because it depends on packet reception. The processed recordings therefore support a small-scale dataset for evaluating exoskeleton-embedded sensing under real wearing conditions and for developing end-to-end models from synchronized human-device signals.
% ORIGINAL END B037
% REVISION BEGIN B037
The human-worn records show that the shared \rev{end-effector} sensing chain preserves the main hip-motion pattern in both use conditions. In Fig.~\ref{fig:human_recording}(a), encoder-based hip angles remain aligned with the myoMOTION reference, with calibrated RMSE values of $5.54 \pm 2.40^\circ$ for bench-mounted walking and $3.19 \pm 1.10^\circ$ for backpack-mounted walking; the residual \rev{differences} \rev{can} \rev{arise from joint-axis misalignment,} soft-tissue motion, and wearable-interface compliance. \rev{Differences in the indoor and outdoor measurement environments may also contribute.} After \rev{baseline} subtraction, the mean absolute \rev{end-effector} torque was $1.05 \pm 0.32$ N\,m in the bench-mounted \rev{admittance-based motion tracking} condition and $0.68 \pm 0.35$ N\,m in the backpack-mounted unpowered condition, indicating low apparent resistance in both walking modes. This low-resistance condition is useful for intention-sensing extensions and wearable data collection, because \rev{the} \rev{low} \rev{measured} \rev{interaction} \rev{torque} \rev{is} \rev{consistent with} limited mechanical \rev{resistance while the end-effector sensors continue recording during walking.} Fig.~\ref{fig:human_recording}(b) further indicates that \rev{end-effector} encoder and torque channels provide stable high-rate records, whereas the wireless IMU channel is more variable because it depends on packet reception. The processed recordings therefore support a small-scale dataset for evaluating exoskeleton-embedded sensing under real wearing conditions and for developing end-to-end models from synchronized human-device signals.
% REVISION END B037

%%%%%%%%%%%%%%%%%%%%%%%%%%%%%%%%%%%%%%%%%%%%%%%%%%%%%%%%%%%%%%%%%%%%%%%%%%%%%%%%
\section{DISCUSSION}

% ORIGINAL BEGIN B038 (baseline lines 249-249)
% Table~\ref{tab:comparison} compares representative rigid or quasi-rigid hip exoskeleton systems using peak torque-to-wearable-mass ratio, sensing configuration, and platform type. Within this scope, the response-adjusted bench-mounted value is higher than the listed benchmarks, and the backpack-mounted value remains comparable with or above several portable systems. More importantly, the proposed system is the only listed platform that combines terminal torque sensing, encoder-based joint measurement, wireless IMUs, and bench-/backpack-mounted operation. This combination supports the central contribution: laboratory testing, drive switching, and mobile validation can share the same wearable structure, quick-release interface, and sensing path. The bench-mounted path further provides stable power, sensing expansion, high-bandwidth acquisition, and high-performance computation, which are difficult to fully integrate into a lightweight backpack-mounted system.
% ORIGINAL END B038
% REVISION BEGIN B038
Table~\ref{tab:comparison} compares representative rigid or quasi-rigid hip exoskeleton systems using peak torque-to-wearable-mass ratio, sensing configuration, and platform type. \rev{For} the \rev{proposed} \rev{platform,} \rev{Table~\ref{tab:comparison}} \rev{reports} \rev{transmission-adjusted} \rev{component-peak estimates normalized by} the \rev{shared} \rev{wearable-structure} \rev{mass, characterizing} the \rev{output} \rev{supported} \rev{by} \rev{the} \rev{same} \rev{human-side} \rev{structure} \rev{under} \rev{either} \rev{actuation configuration.} More importantly, the proposed system is the only listed platform that combines \rev{end-effector} torque sensing, encoder-based joint measurement, wireless IMUs, and bench-/backpack-mounted operation. This combination supports the central contribution: laboratory testing, \rev{actuation} switching, and mobile validation can share the same wearable structure, \rev{quick-release/assembly} interface, and sensing path. The bench-mounted \rev{configuration} further provides stable power, sensing expansion, high-bandwidth acquisition, and high-performance computation, which are difficult to fully integrate into a lightweight backpack-mounted system.
% REVISION END B038

\begin{table}[t]
\caption{Comparison with Representative Hip Exoskeleton Systems}
\label{tab:comparison}
\centering
\scriptsize
\setlength{\tabcolsep}{3.8pt}
\renewcommand{\arraystretch}{0.92}
\begin{tabular}{@{}lccc@{}}
\toprule
System & TWR & $S_e$ & Platform \\
\midrule
Stanford \cite{Bryan2021} & 14.3 & LC+E & BM \\
BLEEX \cite{Zoss2006,Chu2005,Kazerooni2005} & 3.7 & N.R. & BP \\
Samsung \cite{Seo2016} & 3.9 & I+A & BP \\
Panasonic \cite{John2017} & 2.2 & LC & BP \\
Michigan \cite{Young2017Timing,Young2017Comparison} & 3.4 & LC+G & BM \\
\textbf{Ours} & \textbf{18.4}/10.2$^\dagger$ & \textbf{T+E+W-I} & \textbf{BM+BP} \\
\bottomrule
\end{tabular}
\vspace{0.2ex}
\begin{flushleft}
\scriptsize
% ORIGINAL BEGIN B039 (baseline lines 273-273)
% TWR: peak torque-to-wearable-mass ratio $\tau/m_w$ (N\,m/kg); $S_e$: sensing. BM/BP: bench-/backpack-mounted; LC/T/E/G/I/A/W-I: load cells/torque sensor/encoder/goniometer/IMU/angle sensor/wireless IMU; N.R.: not reported. $^\dagger$Ours gives BM/BP estimates using $m_w=2.89$ kg and $\bar{\eta}_{\tau}=55.7\%$. Bold highlights our system, its sensing and platform features, and the highest listed TWR.
% ORIGINAL END B039
% REVISION BEGIN B039
TWR: peak torque-to-wearable-mass ratio $\tau/m_w$ (N\,m/kg); $S_e$: sensing. BM/BP: bench-/backpack-mounted; LC/T/E/G/I/A/W-I: load cells/torque sensor/encoder/goniometer/IMU/angle sensor/wireless IMU; N.R.: not reported. $^\dagger$\rev{Ours:} BM/BP \rev{component-peak} estimates using \rev{a common assumed transfer ratio of 55.7\% and the 2.89} kg \rev{shared wearable structure, excluding actuation, computing hardware,} and \rev{batteries.} Bold highlights our system, its sensing and platform features, and \rev{our} \rev{higher} \rev{TWR} \rev{estimate.}
% REVISION END B039
\end{flushleft}
\end{table}

% ORIGINAL BEGIN B040 (baseline lines 277-277)
% The backpack-mounted configuration currently uses open-loop torque control. Its terminal response ratios were $56.7 \pm 1.7\%$ during ramp-up, $58.4 \pm 2.0\%$ during holding, and $54.7 \pm 2.7\%$ for sinusoidal amplitude tracking, indicating transmission loss alongside a repeatable torque-transfer relationship. These losses are consistent with the effects of friction, cable-path curvature, hysteresis, and pretension in Bowden transmissions \cite{Jammot2025,Chen2014}. Future work will optimize pretension and cable routing to reduce mechanical losses and use the torque sensors already integrated into the end-effectors to implement closed-loop terminal torque control. This feedback will be used to compensate for the effects of transmission loss and hysteresis on torque tracking and improve output consistency across cable configurations.
% ORIGINAL END B040
% REVISION BEGIN B040
The backpack-mounted configuration currently uses open-loop torque control. \rev{The} \rev{similar} \rev{torque-transfer} ratios \rev{across} ramp-up, holding, and sinusoidal \rev{loading} \rev{indicate} \rev{consistent} \rev{output} \rev{attenuation} \rev{under} \rev{the} \rev{tested} \rev{cable} \rev{configuration.} These losses are consistent with the effects of friction, cable-path curvature, hysteresis, and pretension in Bowden transmissions \cite{Jammot2025,Chen2014}. Future work will optimize pretension and cable routing to reduce mechanical losses and use the torque sensors already integrated into the end-effectors to implement closed-loop \rev{end-effector} torque control. This feedback will be used to compensate for the effects of transmission loss and hysteresis on torque tracking and improve output consistency across cable configurations.
% REVISION END B040

The human-worn results further clarify the remaining system-level limitations. The lower comfort score in the backpack-mounted condition was mainly associated with the larger power supply used to support outdoor recording endurance, rather than with a change in the wearable hip interface. Reducing battery and power-electronics mass, improving load distribution, and refining the waist-back enclosure are therefore important next steps. In parallel, the bench-mounted configuration will be used to implement rehabilitation-oriented algorithms, including impedance training, assistive training, and active training, followed by clinical validation. The backpack-mounted configuration will be extended with gait-phase recognition and end-to-end control algorithms for outdoor assistance studies with more participants and broader terrain conditions.

%%%%%%%%%%%%%%%%%%%%%%%%%%%%%%%%%%%%%%%%%%%%%%%%%%%%%%%%%%%%%%%%%%%%%%%%%%%%%%%%
\section{CONCLUSIONS}

% ORIGINAL BEGIN B041 (baseline lines 284-284)
% This paper presented a reconfigurable bidirectional cable-driven hip exoskeleton platform that allows bench-mounted and backpack-mounted actuation units to drive the same cable-free wearable hip interface. By integrating torque sensing and angle measurement into quick-release end-effectors, and by using wireless IMUs with adaptable electrical access, the platform preserves comparable terminal torque, joint motion, and human motion data in both configurations. Human-worn experiments validated drive replacement, bench-mounted admittance-following response, backpack-mounted open-loop torque response, and continuous sensing records. The platform provides a hardware route for iterative exoskeleton development from laboratory prototyping and rehabilitation training toward mobile assistance validation. Future work will use the bench-mounted configuration for systematic indoor gait-correction and admittance-interaction experiments, and the backpack-mounted configuration for outdoor active-assistance tests that combine gait phase estimation, trajectory planning, torque control, and longer-term wearable evaluation.
% ORIGINAL END B041
% REVISION BEGIN B041
This paper presented a reconfigurable bidirectional cable-driven hip exoskeleton platform that allows bench-mounted and backpack-mounted actuation units to drive the same cable-free wearable hip interface. By integrating torque sensing and angle measurement into \rev{quick-release/assembly} end-effectors, and by using wireless IMUs with \rev{compatible} \rev{sensor} \rev{and controller connections,} the platform preserves comparable \rev{end-effector} torque, joint motion, and human motion data in both \rev{configurations, allowing experiments to reuse the body attachments and measurement arrangement when the actuation unit is replaced.} Human-worn experiments validated \rev{actuation-unit} replacement, bench-mounted \rev{admittance-based motion tracking} response, backpack-mounted open-loop torque \rev{tracking,} and continuous \rev{joint-motion and interaction-torque} records. The platform provides a hardware route for iterative exoskeleton development from laboratory prototyping and rehabilitation training toward mobile assistance validation. Future work will use the bench-mounted configuration for systematic indoor gait-correction and admittance-interaction experiments, and the backpack-mounted configuration for outdoor active-assistance tests that combine gait phase estimation, trajectory planning, torque control, and longer-term wearable evaluation. \rev{This progression will also examine how shared sensing conventions support the adaptation and reuse of higher-level algorithms between laboratory and mobile experiments.}
% REVISION END B041

%%%%%%%%%%%%%%%%%%%%%%%%%%%%%%%%%%%%%%%%%%%%%%%%%%%%%%%%%%%%%%%%%%%%%%%%%%%%%%%%

\end{document}